\documentclass[sigconf]{aamas} 
\usepackage{balance} % for balancing columns on the final page

\setcopyright{ifaamas}
\acmConference[AAMAS '23]{Proc.\@ of the 22nd International Conference
on Autonomous Agents and Multiagent Systems (AAMAS 2023)}{May 29 -- June 2, 2023}
{London, United Kingdom}{A.~Ricci, W.~Yeoh, N.~Agmon, B.~An (eds.)}
\copyrightyear{2023}
\acmYear{2023}
\acmDOI{}
\acmPrice{}
\acmISBN{}

\acmSubmissionID{71}

\title[AAMAS-2023 Formatting Instructions]{Persuading to Prepare for Quitting Smoking with a Virtual Coach: Using States and User Characteristics to Predict Behavior}

\author{Nele Albers}
\affiliation{
  \institution{Delft University of Technology}
  \city{Delft}
  \country{The Netherlands}}
\email{n.albers@tudelft.nl}

\author{Mark A. Neerincx}
\affiliation{
  \institution{Delft University of Technology}
  \city{Delft}
  \country{The Netherlands}}
 \affiliation{
  \institution{TNO}
  \city{Soesterberg}
  \country{The Netherlands}}
\email{m.a.neerincx@tudelft.nl}

\author{Willem-Paul Brinkman}
\affiliation{
  \institution{Delft University of Technology}
  \city{Delft}
  \country{The Netherlands}}
\email{w.p.brinkman@tudelft.nl}

\begin{abstract}
Despite their prevalence in eHealth applications for behavior change, persuasive messages tend to have small effects on behavior. Conditions or states (e.g., confidence, knowledge, motivation) and characteristics (e.g., gender, age, personality) of persuadees are two promising components for more effective algorithms for choosing persuasive messages. However, it is not yet sufficiently clear how well considering these components allows one to predict behavior after persuasive attempts, especially in the long run. Since collecting data for many algorithm components is costly and places a burden on users, a better understanding of the impact of individual components in practice is welcome. This can help to make an informed decision on which components to use. We thus conducted a longitudinal study in which a virtual coach persuaded 671 daily smokers to do preparatory activities for quitting smoking and becoming more physically active, such as envisioning one's desired future self. Based on the collected data, we designed a Reinforcement Learning (RL)-approach that considers current and future states to maximize the effort people spend on their activities. Using this RL-approach, we found, based on leave-one-out cross-validation, that considering states helps to predict both behavior and future states. User characteristics and especially involvement in the activities, on the other hand, only help to predict behavior if used in combination with states rather than alone. We see these results as supporting the use of states and involvement in persuasion algorithms. Our dataset is available online.
\end{abstract}


\newcommand{\BibTeX}{\rm B\kern-.05em{\sc i\kern-.025em b}\kern-.08em\TeX}

\begin{document}

%%% The following commands remove the headers in your paper. For final 
%%% papers, these will be inserted during the pagination process.

\pagestyle{fancy}
\fancyhead{}

%%% The next command prints the information defined in the preamble.

\maketitle 

%%%%%%%%%%%%%%%%%%%%%%%%%%%%%%%%%%%%%%%%%%%%%%%%%%%%%%%%%%%%%%%%%%%%%%%%

\section{Introduction}

Recent years have seen a surge of eHealth applications for behavior change (e.g., \cite{ly2017fully, fadhil2017addressing, meijer2021least}), which provide behavior change support over the Internet or connected technologies such as apps and text messaging. Such applications often ask their users to do activities such as setting a goal, planning a running route, or watching an educational video. Persuasive messages are commonly used to motivate users to do these activities. For example, users may be reminded that doing an activity is in line with their decision to change their behavior. However, the effect of single persuasive attempts on behavior tends to be small (e.g., \cite{kaptein2015personalizing, Vries2018TheoryBasedAT, albers2022addressing}). 

Several studies have tried to increase the effectiveness of a persuasive attempt. One way is to consider the current state people are in (e.g., confidence, knowledge, motivation). Such a state describes a person's condition or status at a certain time that is relatively stable with regards to its elements \cite{apaState}. \citet{carfora2020dialogue} and \citet{klein2013intelligent}, for instance, account for people's self-efficacy when selecting messages for behavior change. Doing so is in line with behavior change theories, which posit that behavior is influenced by people's current state (e.g., \cite{michie2011behaviour, ajzen1991theory}). Yet, behavior in turn can also influence people's states. For example, verbally persuading people \cite{strecher1986role} or improving their mood \cite{kavanagh1985mood} may increase their self-efficacy. Intuitively, we want to persuade people in such a way that they move to a state in which they are more likely to be successfully persuaded again. One framework that allows one to consider both current and future states is Reinforcement Learning (RL). RL with consideration of states has been applied to adapting the framing of messages for inducing healthy nutritional habits \cite{carfora2020dialogue} or the affective behavior of a social robot teacher \cite{gordon2016affective}. However, it is not yet sufficiently clear how persuasive attempts affect behavior and future states, especially after a sequence of these attempts. 

An alternative to considering people's current state when choosing a persuasive strategy is to consider their characteristics such as gender, personality, and involvement in an issue. While previous work has found such characteristics to play a role (e.g., \cite{kaptein2012heterogeneity, Vries2018TheoryBasedAT, maheswaran1990influence}), little work has comprehensively compared the use of user characteristics to the one of states. In addition, it may be helpful to combine these two approaches: behavior after applying a persuasive strategy in a state may differ based on user characteristics.

Our goal thus is to shed light on the effects of considering algorithm components such as states, user characteristics, or both when choosing a persuasive strategy. While previous work has tested algorithms with such components (e.g., \cite{gordon2016affective, hors2019opening}), we do not yet understand the effects of individual algorithm components in practice. Therefore, rather than developing a new algorithm and comparing it to existing ones, we want to first get a better understanding of the practical impact of algorithm components. This can enable informed decisions on which components to include, which is desirable due to the larger amount of human data that needs to be collected when more components are used. Collecting more human data is costly and places a burden on users of eHealth applications that is unlikely to benefit the already low adherence rates to these applications. If data is explicitly collected by means of questions, people are likely to stop using the application if many questions are asked. For example, \citet{pommeranz2012designing} saw that more cognitively demanding preference elicitation methods were seen as more effortful and liked less, which is negatively associated with technology use \cite{venkatesh2012consumer}. Moreover, while implicit data collection methods such as sensors have the potential to collect high-quality data less obtrusively, they often do not yet succeed at this. \citet{yang2023just}, for instance, found in the context of smoking cessation that improvements in sensing technology are needed to obtain higher data quality, lower the burden to users, and increase adherence.

Thus, to get a better understanding of the effects of algorithm components, we conducted a study in which smokers interacted with the text-based virtual coach Sam in up to five sessions. In each session, Sam assigned people a new preparatory activity for quitting smoking together with a persuasive strategy. The goal of these activities was to prepare people for change, which is typically done at the start of a behavior change intervention to increase the likelihood of successful change. Half of the activities targeted becoming more physically active as this may facilitate quitting smoking \cite{haasova2013acute, trimbos2016}. In the next session, Sam asked about the effort people spent on their activity to measure their behavior. To determine people's states, Sam asked questions about people's capability, opportunity, and motivation to do an activity. Each pair of states from consecutive sessions forms a transition sample that we used to predict states after persuasive attempts. Moreover, we measured 32 characteristics covering demographics, smoking and physical activity, personality, and involvement in the activities. Based on the resulting 2366 transition samples from 671 people, we compared the effectiveness of considering states, user characteristics, or both for predicting behavior after persuasive attempts. In addition, we used simulations to assess the long-term effects of optimally persuading people based on an RL-approach that considers current and future states to maximize the effort people spend on their activities. 

This paper's contribution is evidence supporting the use of states derived from behavior change theories as well as people's overall involvement as components in persuasion algorithms. Following the stages in the development of technological health interventions defined by \citet{brinkman2011guest}, this justifies research on including these components in a full intervention as a next step.

\section{Background}

\subsection{Persuasive strategies}

Several sets of persuasive strategies have been defined. For example, \citet{oinas2008systematic} distinguish seven persuasive strategies such as social comparison and competition, \citet{cialdini2006influence} defines six persuasive strategies such as authority, \citet{fogg2002persuasive} differentiates between persuasive strategies related to \textquotedblleft technology as a tool\textquotedblright\space (e.g., self-monitoring) and those related to \textquotedblleft technology as a social actor\textquotedblright\space (e.g., language cues), and \citet{consolvo2009theory} describe nine persuasive strategies such as credibility. Such persuasive strategies are meant to directly influence people's motivation \cite{michie2011behaviour}. In addition, there are strategies that are meant to influence motivation indirectly by, for example, restructuring a person's environment. Examples include action and coping planning \cite{sniehotta2005action}. Notably, many of these persuasive strategies can be implemented in several ways. For instance, there are different ways of framing messages (e.g., \cite{steward2003need, catellani2021connecting}) and communication modalities (e.g., \cite{vidrine2012randomized, wang2019guided}). In this work, we focus on persuasive strategies that can be implemented in a text-based virtual coach that supports a single person in their behavior change process, without requiring external elements such as sensor data or peers. We thereby interpret the term persuasion broadly to also include strategies that influence motivation indirectly.

\subsection{States}

Persuasive strategies are not equally effective in all circumstances: the context of a persuasive attempt matters \cite{alslaity2020impact, oinas2009persuasive}. One way to describe the context is the state a persuadee is in. For example, the effectiveness of different health messages depends on a persuadee's self-efficacy \cite{Bertolotti2019DifferentFT}, and the processing of messages depends on a persuadee's mood \cite{Bless1990MoodAP, fogg2002persuasive}. Several of these state features have been formalized as influencing behavior in behavior change theories. One such theory is the behavior change wheel \cite{michie2011behaviour}, at whose center lies the Capability-Opportunity-Motivation-Behavior (COM-B) model of behavior. This COM-B model is an overarching causal model of behavior, according to which a person's capability, motivation, and opportunity determine their behavior. Capability includes having the necessary knowledge and skills, motivation considers the brain processes influencing behavior, and opportunity captures factors outside of an individual such as support from one's social environment. The COM-B model is overarching in the sense that components of other behavior change theories can be mapped to it. For example, Fogg's behavior model specifies that ability, motivation, and a trigger need to come together for behavior to happen \cite{fogg2002persuasive}. Ability can be mapped to \textquotedblleft Capability\textquotedblright\space in the COM-B model, motivation to \textquotedblleft Motivation,\textquotedblright\space and the trigger to \textquotedblleft Opportunity.\textquotedblright\space The COM-B model thus provides an indication of which information about a persuadee's state needs to be considered to predict behavior after persuasive attempts. One question we pose hence is:

\begin{center}
\textit{Q1: How well can states derived from the COM-B model predict behavior after persuasive attempts?}
\end{center}

\subsection{Future states}

In the COM-B model, a person's capability, opportunity, and motivation influence their behavior, and the behavior in turn influences their capability, opportunity, and motivation. Thus, behavior influences people's future states. This effect of behavior on a person's state has also been studied in the context of persuasion. For instance, \citet{steward2003need} found that the framing of messages influences their effect on self-efficacy, and \citet{carfora2019informational} saw that the message type affects a person's intention to act, anticipated regret, and attitude towards behavior. Thus, persuasive strategies differ in their effect on a persuadee's state. Ideally, we would choose a persuasive strategy that positively influences a persuadee's state by, for example, increasing motivation. To do so, we need to be able to predict not just the behavior, but also the state after a persuasive attempt. We thus investigate the following question:

\begin{center}
    \textit{Q2: How well can states derived from the COM-B model predict states after persuasive attempts?}
\end{center}

Ideally, a persuasive attempt moves a person to a future state in which they are more likely to be successfully persuaded again. Since capability, opportunity, and motivation determine behavior, the goal is that each person ultimately moves to, and then stays in, a state with high values for these predictors of behavior. We, therefore, want to examine what happens to people's states after a sequence of persuasive attempts in the ideal case. The ideal case is that we always use the optimal persuasive strategy:

\begin{center}
\textit{Q3: What is the effect of (multiple) optimal persuasive attempts on persuadees' states?}
\end{center}

Being able to predict states may help to choose effective sequences of persuasive strategies, but how is behavior affected by using sequences of optimal persuasive strategies? And importantly, how much does it matter what a virtual coach says? Hence, we pose the following question:

\begin{center}
    \textit{Q4: How do optimal and sub-optimal persuasive attempts compare in their effect on behavior?}
\end{center}

\subsection{User characteristics}

Considering people's states is one way to capture their differing responses to persuasive strategies - considering user characteristics is another. With user characteristics we mean information about a user that changes, if at all, very slowly and irrespective of persuasive attempts. \citet{kaptein2012heterogeneity}, for instance, showed that age, gender, and personality may influence which of the persuasive strategies by \citet{cialdini2006influence} is most effective. Several other works have confirmed the influence of user characteristics such as the stage of behavior change \cite{Vries2018TheoryBasedAT}, personality \cite{alkics2015impact, Vries2018TheoryBasedAT, halko2010personality, oyibo2017effects, zalake2021effects}, age and gender \cite{muhammad2018personalizing}, cultural background \cite{oyibo2018susceptibility}, and how people approach pleasure and pain \cite{cesario2008regulatory}. Another potentially important user characteristic is involvement. According to the Elaboration Likelihood Model (ELM) \cite{petty1986elaboration}, messages are more likely to be processed in detail when people are highly involved in an issue \cite{maheswaran1990influence}. Such in-depth processing in turn is more likely to have a persistent effect \cite{petty1986elaboration}. Predicting the effectiveness of persuasive attempts based on user characteristics has the advantage that we need to collect data less often: in contrast to states, we do not need to gather this data before each persuasive attempt. We thus pose the following question:

\begin{center}
    \textit{Q5: How does predicting behavior based on user characteristics compare to doing so based on states?}
\end{center}

Rather than \textit{replacing} states with user characteristics, one may also \textit{use both} states and characteristics. For instance, \citet{steward2003need} showed that a person's need for cognition influences the effect of message types on self-efficacy. Thus, user characteristics may have an effect on the states after persuasive attempts. Intuitively, one would expect people who are more similar with regard to these user characteristics to respond more similarly to persuasive attempts. We, therefore, investigate the following question:

\begin{center}
    \textit{Q6: How does incorporating users' similarity based on characteristics, besides the consideration of states, improve the prediction of behavior?}
\end{center}

\section{Methodology}

To answer our research questions, we developed the virtual coach Sam that persuaded people to do preparatory activities for quitting smoking based on an RL-algorithm. This algorithm for choosing persuasive strategies aimed to maximize the effort people spend on their activities over time. Data for the algorithm was collected in a longitudinal study. The data and analysis code underlying this paper as well as the Appendix can be found online \cite{Albers2023aamasdataset}.

\subsection{Virtual coach}
We implemented the text-based virtual coach Sam that helped people prepare for quitting smoking and becoming more physically active in conversational sessions. In each session, Sam randomly proposed to users a new preparatory activity for quitting smoking or becoming more physically active such as tracking one's smoking behavior. These activities were based on components of the StopAdvisor smoking cessation intervention \cite{michie2012development} and future-self exercises \cite{meijer2018strengthening, penfornis2023my}. After proposing the activity, Sam asked questions to determine a user's current state. This state was used as input for choosing how to persuade the user to do the activity. In the next session, Sam asked about users' experience with their activity and the effort they spent on it. Throughout the dialog, Sam used techniques from motivational interviewing \cite{Henkemans2009AnOL} such as giving compliments for spending a lot of effort on activities and otherwise expressing empathy. Empathy can also facilitate forming and maintaining a relationship with a user \cite{bickmore2005s}, which can support behavior change \cite{zhang2020artificial}. Moreover, based on discussions with smoking cessation experts, Sam maintained a generally positive and encouraging attitude while trying to avoid responses that may be perceived as too enthusiastic \cite{free2009txt2stop}. The implementation of the virtual coach, based on Rasa and Rasa Webchat, can be found online \cite{albers2021virtualcoachcode}. The structure and an example of the conversational sessions as well as examples of the activities are available in the Appendix.

\subsection{Persuasion algorithm}
For each persuasive attempt, Sam chose a persuasive strategy based on its learned policy. In the next session, the user provided Sam with feedback by reporting the effort they spent on their activity. Formally, we can define our approach as a Markov Decision Process (MDP) $\langle S, A, R, T, \gamma \rangle$. The action space $A$ consisted of different persuasive strategies, the reward function $R: S \times A \times S \rightarrow [-1, 1]$ was determined by the self-reported effort, $T: S \times A \times S \rightarrow [0, 1]$ described the transition function, and the discount factor $\gamma$ was set to $0.85$ to favor rewards obtained in the near future over rewards obtained in the more distant future. The intuition behind this value for $\gamma$ was that while we wanted to persuade a user over multiple time steps successfully, a failed persuasive attempt in the near future could cause a user to become less receptive to future ones or even to drop out entirely: early success might encourage people to continue \cite{amabile2011progress}. The finite state space $S$ described the state a user was in and was captured by answers to questions about a user's capability, opportunity, and motivation to perform an activity \cite{michie2014behaviour}. The goal of an agent in an MDP is to learn an optimal policy $\pi^*: S \rightarrow \Pi(A)$ that maximizes the expected cumulative discounted reward $\mathbb{E}\big[\sum_t^\infty \gamma^tr_t\big]$ for acting in the given environment. The value function $V^\pi: S \rightarrow \mathbb{R}$ describes the expected cumulative discounted reward for executing $\pi$ in state $s$ and all subsequent states. $V^*$ denotes the value function if $\pi = \pi^*$. Figure~3 in the Appendix illustrates the algorithm idea.

\paragraph{State space.}
In each session, users provided answers to questions about their capability, opportunity, and motivation to do preparatory activities (e.g., \textquotedblleft I feel that I need to do the activity\textquotedblright) on 5-point Likert scales. These questions were based on the COM-B self-evaluation questionnaire \cite{michie2014behaviour} with an additional question about self-efficacy based on \citet{sniehotta2005long} to assess motivation (see Table~2 in the Appendix). To use the time and effort of users efficiently, we only asked those questions that we envisioned to differ between people for our domain. We transformed the questions to binary features based on whether a value was greater than or equal to the feature mean (1) or less than the feature mean (0). To further reduce the size of the state space, we used our collected data to select three out of eight features in a way that was inspired by the G-algorithm \cite{chapman1991input}. This involved iteratively selecting the feature for which the Q-values were most different when the feature is 0 compared to when the feature is 1. Besides the reduction in state space size, this feature selection also has the benefit that fewer questions would need to be answered by users in practice. The three chosen features were 1) whether users felt like they wanted to do an activity, 2) whether they had things that prompted or reminded them to do an activity, and 3) whether they felt like they needed to do an activity. The resulting state space had a size of $|S| = 2^3 = 8$. We denote states with binary strings such as $001$ (here the first and second features are $0$ and the third feature is $1$).

\paragraph{Action space.}
Five persuasive strategies formed the action space: authority, commitment, and consensus from \citet{cialdini2006influence}, action planning \cite{hagger2014implementation}, and no persuasion. The first three persuasive strategies consisted of a persuasive message (e.g., \textquotedblleft Experts recommend $\langle doing\ activity \rangle$ to $\langle positive \ impact \ of \ activity \rangle$.\textquotedblright) and a subsequent reflective question (e.g., \textquotedblleft Which other experts, whose opinion you value, would agree with this?\textquotedblright). The latter was meant to increase the in-depth central processing of the persuasive message. According to the ELM, such high-effort central processing of messages leads to attitudes that are more likely to be persistent over time, resistant to counterattack, and influential in guiding thought and behavior \cite{petty1986elaboration}. Persuasive messages were based on the validated messages from \citet{thomas2017adapting}. For action planning, users were asked to create an if-then plan for doing their activity based on the formulation by \citet{sniehotta2005long}. Yet, rather than asking users to enter their action plans in a table, the virtual coach prompted them to create an if-then plan of the form \textquotedblleft If $\langle$situation$\rangle$, then I will $\langle$do activity$\rangle$\textquotedblright\space based on \citet{chapman2009comparing}. For the first four persuasive strategies, a message that reminded people of their new activity after the session also contained a question based on the persuasive strategy. These reminder questions were based on the ones by \citet{schwerdtfeger2012using}. Repeating a persuasive attempt can also increase in-depth central processing \cite{petty1986elaboration}. Examples of persuasive messages and reflective questions are given in the Appendix.

\subsubsection*{Reward}

In sessions 2--5, participants were asked about the overall effort they spent on their last activity on a scale from 0 to 10, adapted from \citet{hutchinson2006perceived}. Based on the mean effort $\overline{e}$, the reward $r \in [-1, 1]$ for an effort $e$ was computed as follows:
\begin{center}
    $ r = \begin{cases}
    -1 + \frac{e}{\overline{e}} & if \ e < \overline{e}\\
    1 - \frac{10 - e}{10 - \overline{e}} & if \ e > \overline{e}\\
    0 & otherwise.
    \end{cases}$
\end{center}
The idea behind this reward signal was that an effort that was equal to the mean was awarded a reward of $0$, and that rewards for efforts greater and lower than the mean were each equally spaced.

\subsection{Data collection}
\paragraph{Study.}
We conducted a longitudinal study in which people interacted with Sam in up to five conversational sessions between 20 May 2021 and 30 June 2021. The Human Research Ethics Committee of Delft University of Technology granted ethical approval for the research (Letter of Approval number: 1523). Before the collection of data, the study was preregistered in the Open Science Framework (OSF) \cite{albers_brinkman_2021}. Participants were recruited from the online crowdsourcing platform Prolific. Eligible were people who were contemplating or preparing to quit smoking \cite{diclemente1991process}, smoked tobacco products daily, were fluent in English, were not part of another intervention to quit smoking, had an approval rate of at least 90\% and at least one previous submission on Prolific, and provided informed consent. Participants were persuaded randomly in the first two sessions. Afterward, participants were split into four groups, each of which was persuaded based on a different policy. We provide details on these policies in Table~5 in the Appendix. 760 people started the first session, and 518 people successfully completed session 5 (see Figure~3 in the Appendix). Participant characteristics such as age and education level are shown in Table~6 in the Appendix.

\paragraph{Data.}
We gathered 2366 $\langle s, a, r, s'\rangle$-samples from 671 people, where $s$ is the state, $a$ the action, $r$ the reward, and $s'$ the next state. Besides these transition samples, we also collected data on user characteristics. This includes 31 pre-characteristics (i.e., characteristics measured before any persuasive attempt) covering demographics, smoking, physical activity, personality, and need for cognition. Moreover, we measured users' overall involvement in their assigned activities after the five sessions. Due to dropout, we obtained involvement data for only 500 participants. The Appendix provides more information on the user characteristics we measured.

\section{Results}

We now investigate each of our six research questions. For each research question, we first describe our setup, followed by our findings and the resulting answer to the research question.\\

\textbf{Q1: How well can states derived from the COM-B model predict behavior after persuasive attempts?}

\paragraph{Setup.} Knowing the state a persuadee is in may help to predict their behavior after persuading them with different persuasive strategies (i.e., actions). The behavior in our case is the effort people spend on their preparatory activities, which is captured by our reward function. We compared two approaches for predicting the reward: 1) the mean reward per action, and 2) the mean reward per action and state. We used leave-one-out cross-validation for the 671 participants with at least one transition sample to compare the two approaches based on the mean $L_1$-error and its Bayesian 95\% credible interval (CI) \cite{oliphant2006bayesian} per state. In contrast to classical confidence intervals, Bayesian CIs provide information on the most likely values (i.e., a likely range) \cite{hoekstra2014robust}. We regard non-overlapping 95\% CIs as a credible indication that values are different, both for this research question and the subsequent ones. 

\begin{figure}[ht]
  \centering
  \includegraphics[width=0.96\linewidth]{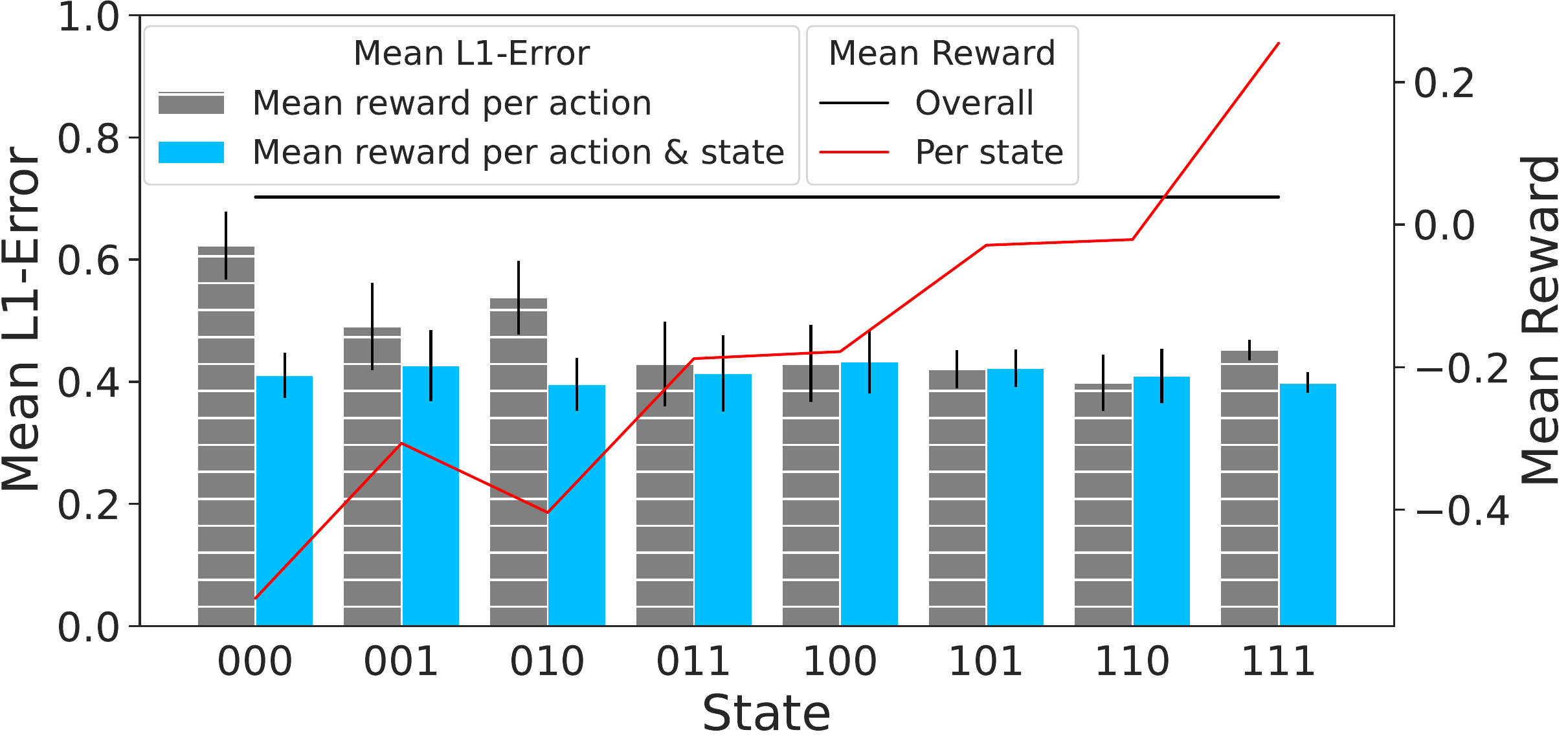}
   \caption{\small Left axis: Mean $L_1$-error with 95\% CIs for predicting rewards based on 1) the mean reward per action and 2) the mean reward per action and state. Right axis: Mean reward overall and per state.}
    \label{fig:rewards}
  \Description{Left axis: Mean L1-error with 95\% credible intervals for predicting rewards based on 1) the mean reward per action and 2) the mean reward per action and state. Right axis: Mean reward overall and per state.}
\end{figure}

\paragraph{Findings.} Considering the state tends to result in lower $L_1$-errors for predicting the reward than not considering the state (Figure~\ref{fig:rewards}). This makes sense, as the mean reward strongly differs between states. For example, while state 000 has a mean reward of -0.52, state 111 has one of 0.25 (see the red line in Figure~\ref{fig:rewards}). In such states with mean rewards much lower or higher than the overall mean reward, the advantage of considering states for the reward prediction is pronounced with the 95\% CIs for the two approaches not overlapping. This provides a credible indication that considering states performs better. For states with mean rewards more similar to the overall mean reward, on the other hand, the 95\% CIs for the two approaches tend to overlap. So there is no credible indication that one of the two approaches is better for those states.

\paragraph{Answer to Q1.} Considering the state a persuadee is in helps to predict the effort they spend on an activity, as long as the state is one in which people tend to spend much less or more effort on activities than on average. Using features derived from the COM-B model, we obtained such states.\\

\textbf{Q2: How well can states derived from the COM-B model predict states after persuasive attempts?}

\paragraph{Setup.} Ideally, we want to persuade a person in such a way that they move to a state in which they are likely to again be persuaded to spend a lot of effort on an activity. Therefore, we need to be able to predict the state after a persuasive attempt. Using leave-one-out cross-validation, we compared three approaches for predicting the next states for the samples from the left-out person: 1) assigning an equal probability to all states, 2) predicting that people stay in their current state, and 3) using the transition function estimated from the training data. We compared the three approaches based on the mean likelihood of the next state and its 95\% CI per state. A higher likelihood suggests that next states can be predicted better.

\begin{figure}[ht]
  \centering
  \includegraphics[width=0.9\linewidth]{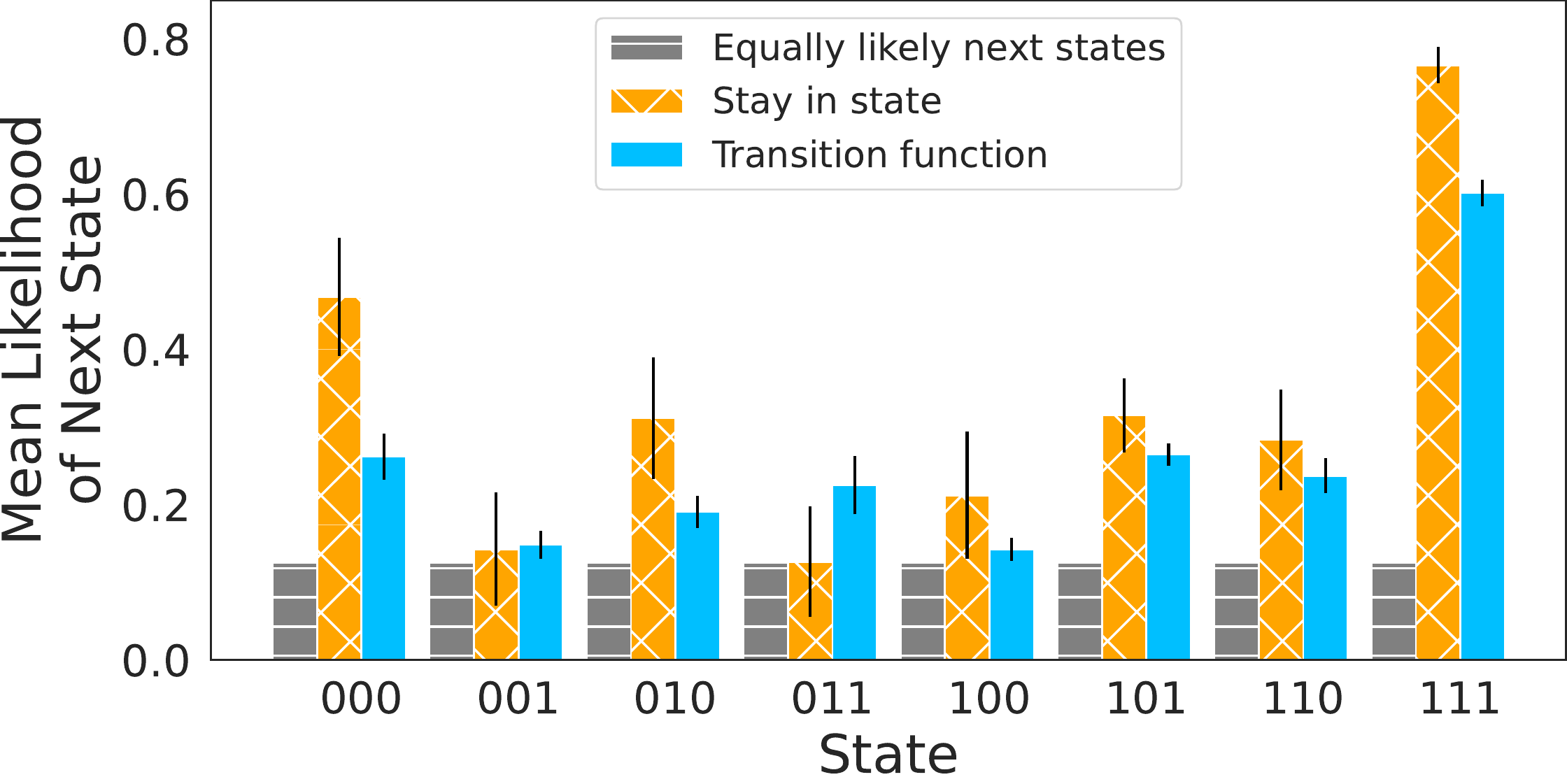}
   \caption{\small Comparison of three approaches to predicting next states with regards to the mean likelihood of next states with 95\% CIs for each state.}
    \label{fig:likelihood_next_states}
  \Description{Bar chart that shows the mean likelihood of next states with 95\% credible intervals for each state for three approaches to predicting next states.}
\end{figure}

\paragraph{Findings.} Figure~\ref{fig:likelihood_next_states} shows that considering the current state, by either predicting that people stay in their current state or assigning a probability to next states based on the estimated transition function, leads to a higher mean likelihood of next states than assigning an equal probability to all next states. This shows that state transitions do not occur uniformly at random. Notably, predicting that people stay in their current state leads to the highest mean likelihood of next states in three of the eight states. These states are states 000, 010, and 111. In each of these states, the mean for predicting that people stay in their current state is highest and the corresponding 95\% CI does not overlap with the ones for the other two approaches. This shows the high probability of staying in those three states, which are states with either very low or very high mean rewards (Figure~\ref{fig:rewards}). 

\paragraph{Answer to Q2.} Our results show that considering the current state a persuadee is in helps to predict their next state after a persuasive attempt. For persuadees who are in states in which people tend to spend very little or very much effort on their activities, this next state tends to be the same as the current one. This means that if we just persuade people as we did in the study used to collect data, we will have limited success in moving people from low-effort to higher-effort states. Though once people are in higher-effort states, they are likely to stay.\\

\textbf{Q3: What is the effect of (multiple) optimal persuasive attempts on persuadees' states?}

\paragraph{Setup.}
We would like that people ultimately move to the states in which they are most likely to be persuaded to spend a lot of effort on activities. Starting from an equal distribution of people across the states, we calculated the percentage of people in each state after following the optimal policy $\pi^*$ for a certain number of time steps. $\pi^*$ was computed via value iteration based on all gathered samples. Table~7 in the Appendix shows $\pi^*$.

\begin{figure}[h]
  \centering
  \includegraphics[width=0.9\linewidth]{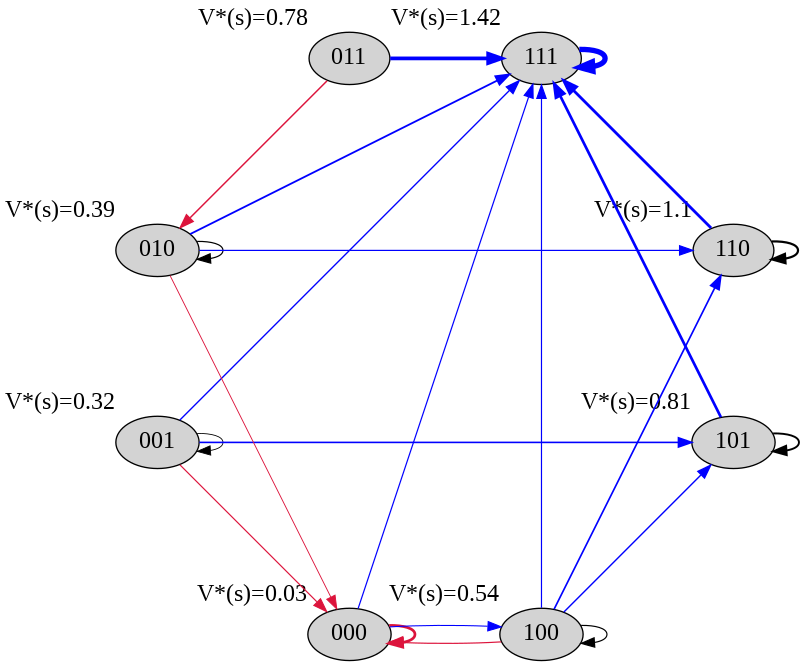}
   \caption{\small Transition probabilities under $\pi^*$. Only transitions with a probability of at least $\frac{1}{|S|}$ are shown. We distinguish transitions to a state with a higher or highest $V^*$ (blue), lower or lowest $V^*$ (red), and the same $V^*$ (black). A thicker line denotes a higher probability.}
    \label{fig:network_graph}
  \Description{Network graph that shows the transition probabilities under the optimal policy.}
\end{figure}

\paragraph{Findings.}
Figure~\ref{fig:network_graph} depicts the transition function under $\pi^*$. It is evident that people tend to move to better states or stay in the best state (blue lines). With better states we mean states with higher $V^*$. In fact, for each state, there is a probability of at least $\frac{1}{|S|}$ that a person moves to a better state. And once people have reached the best state, which is state 111, there is a high probability of 0.8 that they stay there. However, there are some red lines in Figure~\ref{fig:network_graph} as well. These lines show that people sometimes move to worse states or stay in the worst state after being persuaded based on $\pi^*$. This happens especially for states with lower $V^*$ such as states 000 and 010. For both of these states, there is also a relatively high probability of staying in them. For example, there is a probability of 0.41 that people stay in state 000 once there. Yet, people can also move from states with relatively high $V^*$ to states with low $V^*$. For state 011, for instance, there is a probability of 0.22 that people move to the lower-value state 010.

\begin{figure}[ht]
  \centering
  \includegraphics[width=0.9\linewidth]{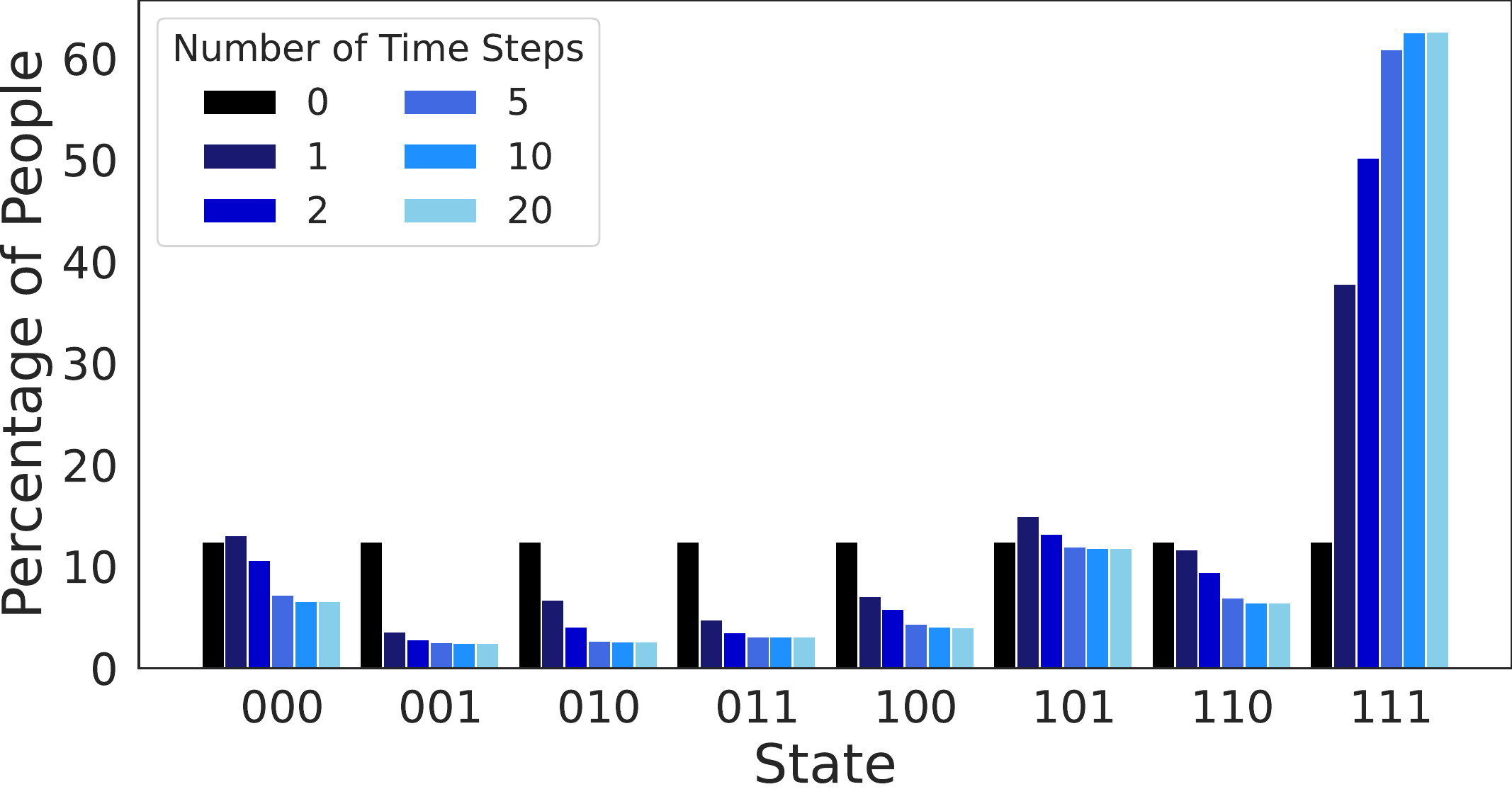}
   \caption{\small Percentage of people in each state after following $\pi^*$ for varying numbers of time steps.}
    \label{fig:states_over_time_opt_policy}
  \Description{Bar chart that shows the percentage of people in each state after following the optimal policy for varying numbers of time steps.}
\end{figure}

Besides the short-term effects of following $\pi^*$, we are also interested in the long-term effects when using multiple persuasive attempts. The results of simulating transitions for applying $\pi^*$ for up to 20 time steps are shown in Figure~\ref{fig:states_over_time_opt_policy}. It is evident that compared to the initial state distribution with an equal number of people in each state, more people are in state 111 and fewer people in all other states after 20 time steps. Given that state 111 is the state with the highest value, people thus tend to move to the best state. In fact, 62.61\% of people and thus more than half are in state 111 after 20 time steps. However, there are always some people in the states with lower values. For example, 6.63\% of people are in state 000, the state with the lowest value, after 20 time steps.

\paragraph{Answer to Q3.} While persuading people optimally multiple times allows most of them to move to and stay in the state in which they are expected to spend the \textit{most} effort on their activities, a few people remain in the state in which they are expected to spend the \textit{least} effort on their activities.\\

\textbf{Q4: How do optimal and sub-optimal persuasive attempts compare in their effect on behavior?}

\paragraph{Setup.}
Once we are able to predict states, we would like to choose effective sequences of persuasive strategies. Yet, it is not clear how much the choice of persuasive strategy matters when it comes to the effort people spend on their activities over time. We calculated the mean reward per transition over time when following 1) the optimal policy $\pi^*$, 2) the worst policy $\pi^-$, and 3) the average policy $\pi^\sim$. $\pi^\sim$ is a theoretical policy for comparison purposes in which each action is taken $\frac{1}{|A|}$ times for each person at each time step, where $|A|$ is the number of actions. We considered two initial state distributions, namely, the distributions across states in the first session of our study based on a) all people and b) only those people whose first reward was in the lowest 25\%-percentile of all first rewards. Distributions are from our study's first session to represent a general population of people who have never been persuaded to do preparatory activities. We further specifically look at people who initially spend very little effort on their activities when persuaded randomly as at the start of our study, because it is more beneficial to coach people who are not yet performing well.

\begin{figure}[ht]
  \centering
  \includegraphics[width=0.96\linewidth]{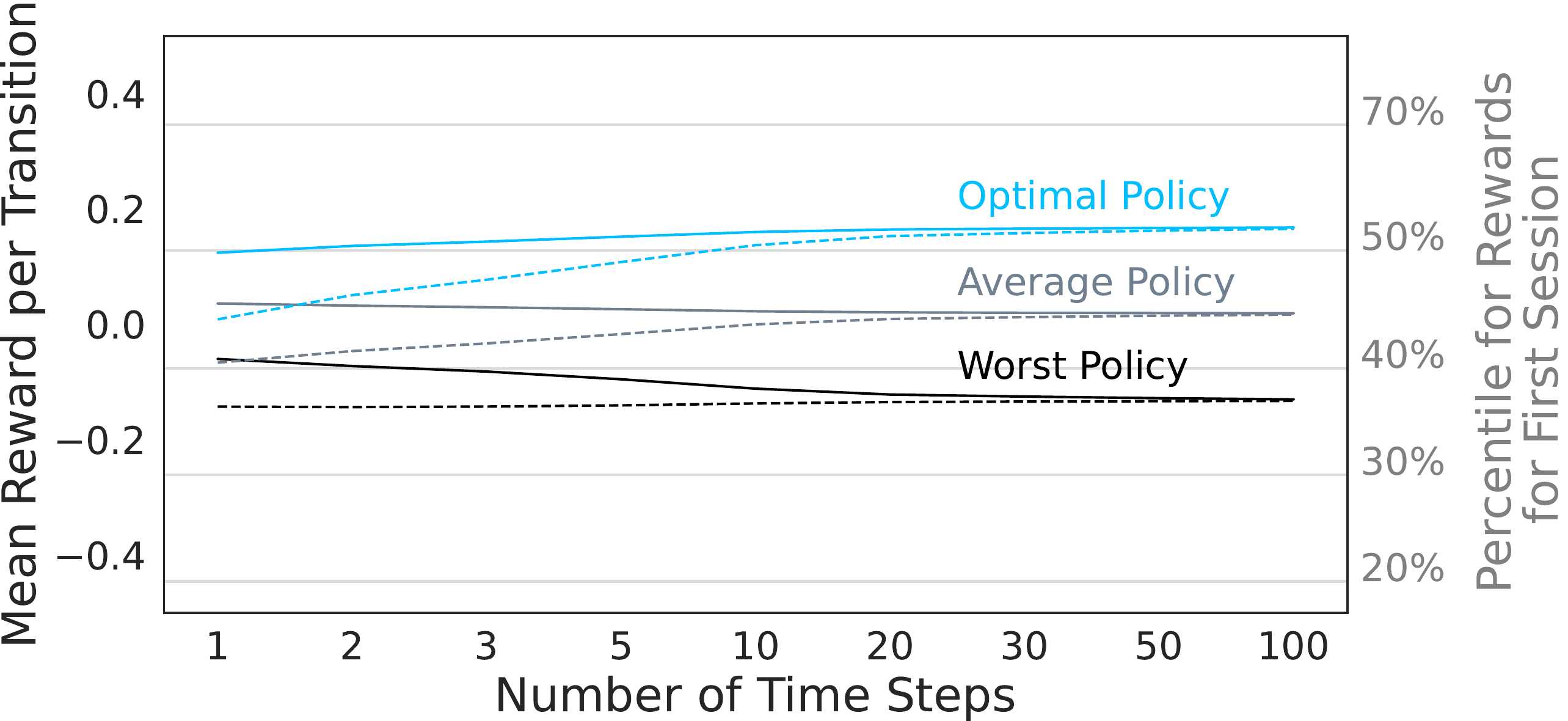}
   \caption{\small Mean reward per transition over time for three policies. The initial populations are the state distribution of all people (solid line) or of only the people with a reward in the lowest 25\%-percentile (dashed line) for the first session.}
    \label{fig:reward_over_time_policy_comparison}
  \Description{Line graph that shows the mean reward per transition over time for three policies. The initial populations are the state distribution of all people (solid line) or of only the people with a reward in the lowest 25\%-percentile (dashed line) for the first session.}
\end{figure}

\paragraph{Findings.}
The mean reward for $\pi^*$ is highest for all time steps and increases over time for an initial state distribution that is based on all people (Figure~\ref{fig:reward_over_time_policy_comparison}). After 100 time steps, the mean reward per transition is 0.17 and therewith above the 50\%-percentile of rewards for the first session of 0.13. This means that the mean reward is increased compared to the actual mean reward we observed in session 1. In contrast, the mean reward drops for the other two policies and is only 0.02 for $\pi^\sim$ and -0.13 for $\pi^-$ after 100 time steps. The former falls in the 40--50\%-percentile of rewards for the first session and the latter in the 30--40\%-percentile. Hence, the difference in mean reward between the three policies increases over time. We also observe this if we consider the initial state distribution of only those people with low rewards for the first session. For example, the difference between $\pi^*$ and $\pi^\sim$ increases from 0.08 to 0.15 and thus almost doubles.

\paragraph{Answer to Q4.} These findings show that it matters, both for people overall and for people who are not performing well initially, how we persuade them to do preparatory activities for quitting smoking. Choosing how to persuade people based on an optimal RL-policy thereby performs better than doing so based on a worst or an average RL-policy.\\

\textbf{Q5: How does predicting behavior based on user characteristics compare to doing so based on states?}

\paragraph{Setup.} An alternative to using states to predict behavior is using user characteristics. This alternative has the advantage that data on such characteristics do not need to be collected before each persuasive attempt. To compare the use of user characteristics to the one of states, we selected three user characteristics in a similar fashion as the three state features. More precisely, we first turned the user characteristics into binary variables based on whether their value was greater than or equal to the mean (1) or less than the mean (0). Then we iteratively selected the variable with the largest difference in reward when the variable is 0 compared to when it is 1. This is because when the reward is very similar for both values of a variable, it does not improve the reward prediction very much to consider the value of the variable. We considered two different sets of candidate variables. First, we considered only the pre-characteristics and thus data that we can collect from people without having to provide any information about the activities (i.e., we excluded people's involvement in their activities). Second, we considered all characteristics (i.e., we also included involvement). The selected characteristics in the first case were the Transtheoretical Model (TTM)-stage for becoming physically active, conscientiousness, and smoking status; the ones in the second case were involvement, physical activity identity, and smoking status. For each case, we created a user characteristic state space of size $2^3 = 8$ analogously to the case of state features. Based on these state spaces, we computed the mean $L_1$-error for predicting the reward using leave-one-out cross-validation. Our baselines were predicting the reward based on 1) the overall mean reward, 2) the mean reward per action, 3) and the mean reward per action and state.

\paragraph{Findings.}

\begin{figure}[ht]
  \centering
  \includegraphics[width=0.9\linewidth]{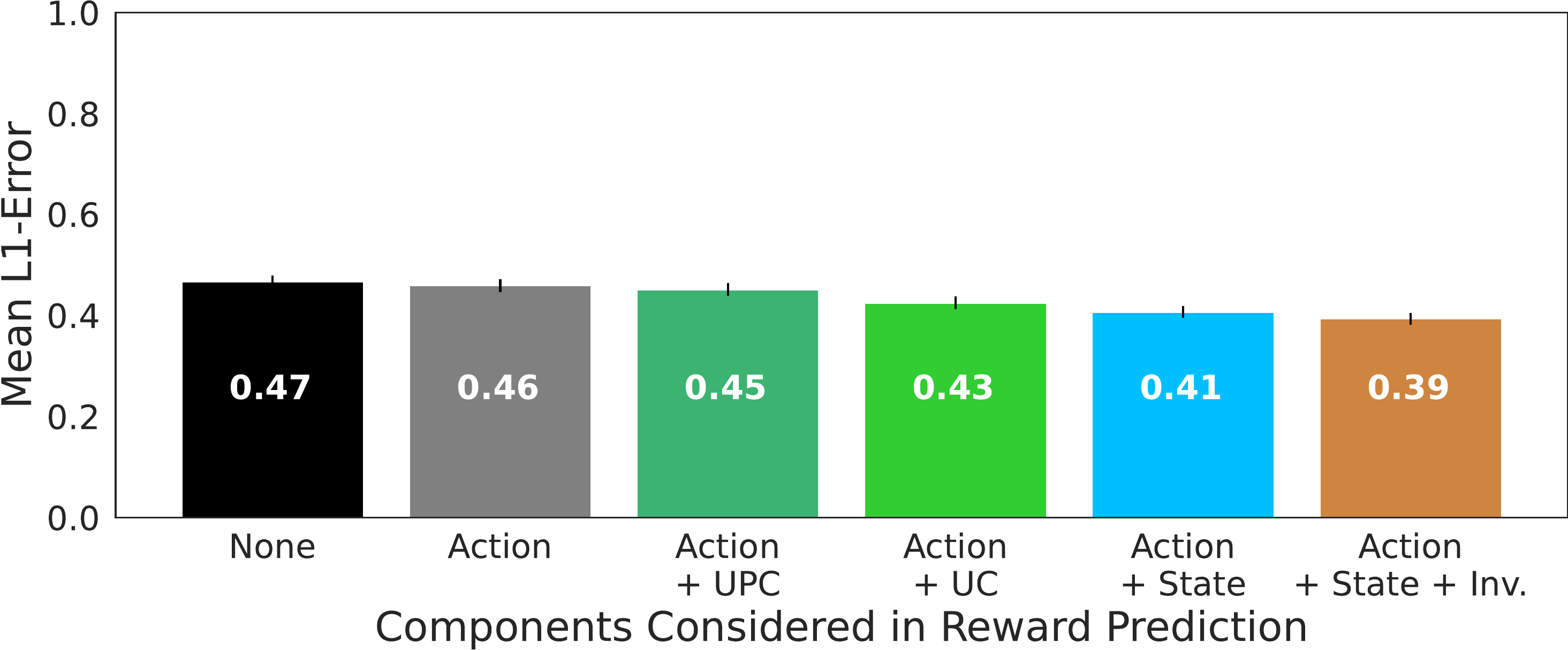}
   \caption{\small Mean $L_1$-error for predicting the reward with 95\% CIs when considering different components for the reward prediction. \textit{None} denotes that we predicted the reward based on the mean overall reward. Abbreviations: UPC, User pre-characteristic; UC, User characteristic; Inv., Involvement.}
    \label{fig:similarity_components_reward}
  \Description{Bar chart that shows the mean L1-error for predicting the reward with 95\% credible intervals when considering different components for the reward prediction.}
\end{figure}

Figure~\ref{fig:similarity_components_reward} shows that predicting rewards based on user characteristics in addition to actions outperforms predicting the overall mean reward. Of the two ways of predicting rewards based on user characteristics, the one that includes people's involvement in their assigned activities leads to a lower $L_1$-error. More precisely, the mean $L_1$-error is 0.43 for user characteristics with involvement, and 0.45 when excluding involvement. The two 95\% CIs thereby do not overlap, providing a credible indication that the mean $L_1$-error is lower for the former than for the latter. However, none of the two ways of predicting rewards based on user characteristics performs better than using states, with the latter leading to a mean $L_1$-error of 0.41. While the 95\% CI for predicting rewards based on states overlaps with the one for predicting rewards based on user characteristics including involvement, it does not overlap with the one for using only user pre-characteristics.

\paragraph{Answer to Q5.} These results provide a credible indication that using states allows us to better predict the effort people spend on their activities than using only user characteristics that we can collect data on without having to tell people about the activities. If we include people's involvement in the activities as a user characteristic, however, there is no longer a credible indication that using states outperforms using user characteristics.\\

\textbf{Q6: How does incorporating users' similarity based on characteristics, besides the consideration of states, improve the prediction of behavior?}

\paragraph{Setup.} While user characteristics \textit{alone} may not help to predict behavior compared to states, they may do so \textit{in combination with states}: people with different characteristics may respond differently to a persuasive attempt in a certain state. We thus examine the effect of incorporating people's similarity, based on user characteristics, on our ability to predict the effort people spend on their activities. We do so by weighting observed samples differently for each persuadee, whereby a larger weight is given to samples from people more similar to the persuadee. Using different user characteristics and weighting parameters, we tried a total of 68 configurations for weighting samples based on similarity (see Appendix). We here report the results for the configuration with the lowest mean $L_1$-error based on leave-one-out cross-validation. This best configuration used people's involvement in their activities to measure similarity.

\paragraph{Findings.} Even though the mean $L_1$-error is lower for incorporating users' similarity than for the original approach without similarity, the 95\% CIs overlap (see the two rightmost bars in Figure~\ref{fig:similarity_components_reward}).

\paragraph{Answer to Q6.} Incorporating users' similarity besides the consideration of states appears to offer some improvement, but there is no credible indication that it allows us to better predict the effort users spend on preparatory activities after persuasive attempts.

\section{Discussion and Conclusion}

The presented study examined the use of states and user characteristics to predict the effort people spend on preparatory activities for quitting smoking after being persuaded by a virtual coach. States were based on the COM-B model and captured people's capability, opportunity, and motivation to do an activity. Our findings suggest that states derived from the COM-B model help to predict behavior: the effort people spend on their activities clearly differs between states (\textit{Q1}). In addition, considering states also helps to predict next states (\textit{Q2}). This may aid in choosing persuasive strategies that move people to future states in which they are more likely to be successfully persuaded again to spend a lot of effort on their activities. With regards to long-term effects, we find based on simulations that people tend to move to better states or stay in the best state when they are persuaded optimally (\textit{Q3}). With good states we mean states in which people are expected to spend a high amount of effort over time when persuaded optimally. However, some people are always in states in which little effort tends to be spent on activities. Our simulation further shows that it matters how we persuade people (\textit{Q4}). More precisely, people tend to spend more effort on activities if they are persuaded optimally based on an RL-algorithm compared to being persuaded based on the worst or an average persuasive strategy. The difference in mean effort per persuasive attempt between the three strategies increases as more persuasive attempts are made before ultimately plateauing. 

Using user characteristics to predict behavior did not perform as well in this study. Compared to using states, we observed worse results when using user pre-characteristics alone (\textit{Q5}). This is the case even though we performed experiments with 31 pre-characteristics that capture a wide range of information about demographics, smoking and physical activity, personality, and need for cognition. Additionally considering users' overall involvement in their activities led to slightly better predictions than considering pre-characteristics alone, but the predictions were still not better than for states. In line with findings by \citet{kaptein2018customizing} in the context of persuasive marketing messages, this suggests that predictions of behavior improve if the predictors are conceptually closer to the behavior. While pre-characteristics such as quitter self-identity may say something about the effort a person is willing to spend to prepare to quit smoking, the person's involvement in such activities is conceptually closer. And states derived from the COM-B model are even closer: they specify theoretically grounded predictors of behavior before each activity. Notably, we find that considering user characteristics \textit{in addition} to states does offer some benefit (\textit{Q6}). But even here, characteristics that are conceptually closer to the behavior we want to predict are most useful, with involvement performing best. However, it may not always be clear how to measure such conceptually closer characteristics. Involvement in our study was, for example, only measured after the persuasive attempts and could thus not inform the selection of persuasive attempts. Asking people to rate prototypes of activities in advance may be a way to address this. As involvement can change, it could also be measured in each session.

\paragraph{Limitations and directions for future work.}
The main limitation of our work is the data it is based on. While we did gather data from human subjects, we did not assess the effects of our approaches on the actual behavior or states of these humans. Instead, we performed leave-one-out cross-validation and simulations. The primary reason is that this allowed us to test a large number of approaches while staying within a reasonable budget. The best-performing approaches can then be tested in the wild in the future. When doing so, however, several additional factors may need to be addressed. This is because all of our approaches assume the transitions between states and the effort people spend on their activities to be stationary. Stationary here means that the transition probabilities and the mean effort people spend for combinations of states and actions do not change. But intuitively, such changes may occur. For instance, repeatedly sending the same persuasive strategy may make it less effective \cite{thomas2017adapting}, but could also help to strengthen the link between cue and response for action planning \cite{schwerdtfeger2012using} or to scrutinize arguments objectively \cite{cacioppo1985central}. One approach to address the effects of such repetitions is the work by \citet{mintz2020nonstationary} on non-stationary bandits. Moreover, once people move beyond preparatory activities and start to actually change their behavior, habits may form after several weeks \cite{gardner2019habit}. Such habits may reduce the cognitive effort and awareness required to do a behavior \cite{gardner2019habit}. One could address this by including information on habits in the state description (e.g., \cite{zhang2022theory}).

% problem of evaluating how good a policy is using off-policy data. See page 63 in phD thesis "Reinforcement learning in models of adaptive medical treatment strategies" by Robert Vincent. Point is that rather than evaluating how good a policy is, we want to see how well we can predict rewards. So we are not really evaluating a policy, but checking how well the Q-values fit. Standard approach mentioned in the thesis is to use only samples that match the new policy. But that would have been very few in our case.

A more general limitation of our work is the way we defined our problem. First, our state description is based on the COM-B self-evaluation questionnaire and only a subset of the questions therein. While this is a good starting point as our results show, other features, potentially derived from other theories, could be useful. For example, physical capability may play a role when people are to be persuaded to do more complex tasks such as going for a run. Importantly, however, not all people may be willing to answer many questions in each session. So it may be beneficial to either limit the number of questions for all people, or to give people the option to answer additional questions for more precise tailoring (e.g., \cite{hors2019opening}). Second, our results are based on five widely used persuasive strategies that we deemed to be applicable in our context. Given the large number of other strategies, it is possible that user characteristics play a more important role in explaining the effectiveness of other strategies. Notably, however, there is also ample literature suggesting the importance of user characteristics for the persuasive strategies we used (e.g., \cite{oyibo2017effects, thomas2017adapting, zalake2021effects}). Third, we measured people's response to persuasive attempts based on the self-reported effort they spent on their activities. It would be interesting to see whether our findings also hold when a more objective measure of behavior is used. Lastly, another interesting direction to improve our model is to use Bayesian RL. This allows one to incorporate prior information about the dynamics in a flexible manner as well as to consider the uncertainty in the learned parameters when making decisions \cite{ross2008model, ghavamzadeh2015bayesian}. For example, one can model relations between state features using a dynamic Bayesian network \cite{ross2008model}. This may be useful, as behavior models such as COM-B specify relations between predictors of behavior.

\paragraph{Conclusion.}
We want to make informed decisions on which components to use in persuasion algorithms for eHealth applications for behavior change that are effective as well as more cost-effective and user-friendly by reducing the amount of required human data. Therefore, a better understanding of the components' individual effects on predicting behavior after persuasive attempts is welcome. We have thus compared the use of states and user characteristics, and a combination thereof, in predicting behavior after persuasive attempts in the context of preparing for quitting smoking with a virtual coach. Our results lend support to the idea of considering states and the user characteristic \textquotedblleft involvement\textquotedblright\space in persuasion algorithms for behavior change. Research on smoking cessation can directly build on these insights and examine the use of these components in a full application. Moreover, both components seem to be domain-independent measures that could also be used in eHealth applications for other behaviors.

%%%%%%%%%%%%%%%%%%%%%%%%%%%%%%%%%%%%%%%%%%%%%%%%%%%%%%%%%%%%%%%%%%%%%%%%

%%% The acknowledgments section is defined using the "acks" environment
%%% (rather than an unnumbered section). The use of this environment 
%%% ensures the proper identification of the section in the article 
%%% metadata as well as the consistent spelling of the heading.

\begin{acks}
This work is part of the multidisciplinary research project Perfect Fit, which is supported by several funders organized by the Netherlands Organization for Scientific Research (NWO), program Commit2Data - Big Data \& Health (project number 628.011.211). Besides NWO, the funders include the Netherlands Organisation for Health Research and Development (ZonMw), Hartstichting, the Ministry of Health, Welfare and Sport (VWS), Health Holland, and the Netherlands eScience Center. The authors acknowledge the help they received from Eline Meijer in formulating the preparatory activities and from Mitchell Kesteloo in hosting the virtual coach on a server. The authors further thank the three anonymous reviewers for their helpful suggestions.
\end{acks}

%%%%%%%%%%%%%%%%%%%%%%%%%%%%%%%%%%%%%%%%%%%%%%%%%%%%%%%%%%%%%%%%%%%%%%%%

%%% The next two lines define, first, the bibliography style to be 
%%% applied, and, second, the bibliography file to be used.

\bibliographystyle{ACM-Reference-Format} 
\balance
\bibliography{bib}

%%%%%%%%%%%%%%%%%%%%%%%%%%%%%%%%%%%%%%%%%%%%%%%%%%%%%%%%%%%%%%%%%%%%%%%%

\end{document}